\documentclass{article}

\usepackage{arxiv}

\usepackage[utf8]{inputenc} 
\usepackage[T1]{fontenc}    
\usepackage{hyperref}       
\usepackage{url}            
\usepackage{booktabs}       
\usepackage{amsfonts}       
\usepackage{nicefrac}       
\usepackage{graphicx} 
\usepackage{microtype}      
\usepackage{lipsum}
\usepackage{mathtools}

\title{Graph Neural Networks to Predict Customer Satisfaction Following Interactions with a Corporate Call Center}

\author{ Teja Kanchinadam, \textsuperscript{\rm 1} Zihang Meng, \textsuperscript{\rm 2} Joseph Bockhorst, \textsuperscript{\rm 1} 
Vikas Singh, \textsuperscript{\rm 2} Glenn Fung \textsuperscript{\rm 1} \\
\textsuperscript{\rm 1} American Family Insurance, Machine Learning Research Group\\
\textsuperscript{\rm 2} University of Wisconsin - Madison \\
\{tkanchin, jbockhor, gfung\}@amfam.com, 
\{vsingh, zmeng29\}@wisc.edu \\
}

\begin{document}
\maketitle

\begin{abstract}
Customer satisfaction is an important factor in creating and maintaining long-term relationships with customers. Near real-time identification of potentially dissatisfied customers following phone calls can provide organizations the opportunity to take meaningful interventions and to foster ongoing customer satisfaction and loyalty. This work describes a fully operational system we have developed at a large US company for predicting customer satisfaction following incoming phone calls. 
The system takes as an input speech-to-text transcriptions of calls and predicts call satisfaction reported by customers on post-call surveys (scale from $1$ to $10$).  Because of its ordinal, subjective, and often highly-skewed nature, predicting survey scores is not a trivial task and presents several modeling challenges.
We introduce a graph neural network (GNN) approach that takes into account the comparative nature of the problem by considering the relative scores among batches, instead of only pairs of calls when training.  This approach produces more accurate predictions than previous approaches including standard regression and classification models that directly fit the survey scores with call data.  Our proposed approach can be easily generalized to other customer satisfaction prediction problems. 
\end{abstract}

\section{Introduction}
\label{sec:introduction}
Improving customer loyalty and tenure are crucial components of business success.
As there is an industry-wide consensus that customer satisfaction is significantly correlated to customer attrition~\cite{custsat},
understanding and improving customer satisfaction are core elements of the strategy of many organizations. 

One aspect of understanding overall customer satisfaction is to monitor and study its determinants within individual interactions, such as phone calls, that customers have with the organization.  
Indeed many businesses conduct automated and / or live customer satisfaction surveys following interactions with corporate call centers. However, due to extremely low response rates, especially for automated surveys, along with the high costs of live surveys, typically only a small percentage of calls have satisfaction scores. Moreover, live calls often have a significant time lag (a week or more) making it difficult to quickly monitor, recognize and respond to both trends and individual cases. In this paper we investigate an approach for shedding light on these unobserved interactions using speech-to-text models and supervised learning.




In previous work we have found that predictions of customer satisfactions derived from simple linear pairwise-ranking models to be a better fit than standard regression methods~\cite{bockhorst2017predicting}, likely due to the ordinal scale of the target variable (the survey responses). We hypothesize, however, that as the relationship between call text and satisfaction is complex and likely non-linear, appropriate non-linear models will lead to more accurate predictions. Here we adapt an approach based on graph neural networks (GNN)~\cite{meng2018efficient} that has previously been found to be effective for non-linear ranking problems.

\begin{figure}
	\centering
		\includegraphics[scale=0.3]{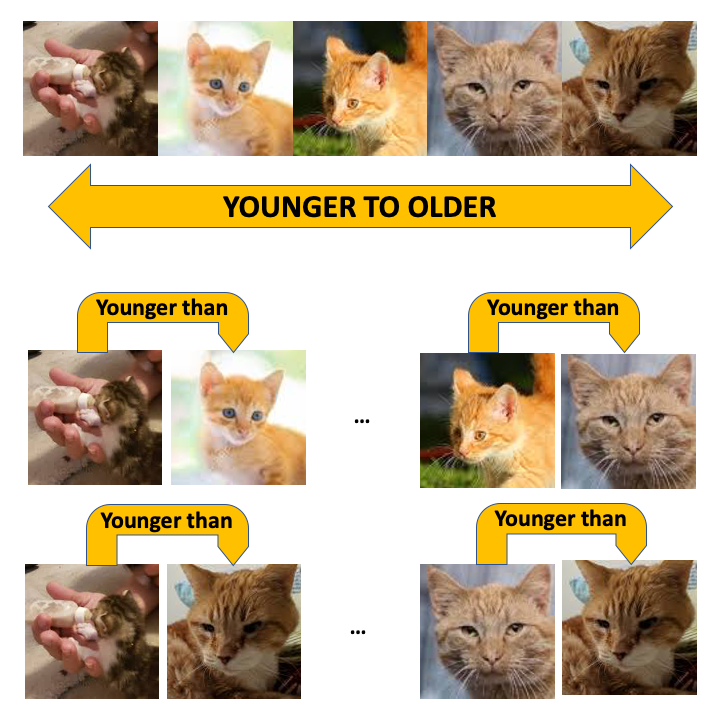}
	\caption{Example - if we are learning to rank cats by age, looking at groups of cats ordered by age simultaneously will provide a more compact representation of the information to the learner than having to learn by only looking at a pair of cats at one time.}
	\label{fig:cats}
\end{figure}

An advantage of the GNN formulation over non-linear pairwise ranking models is that under the GNN predicted rankings for a batch of examples is guaranteed to be coherent with respect to the transitive nature of a ranking. A non-linear pairwise model, on the other hand, has no structural constraint preventing an incoherent set of predictions such as $a > b$, $b > c$ and $c > a$.
Additionally, the GNN may consider higher order features ({\em ie}, non-pairwise) among the examples.
For example, if we are learning to rank cats by age (Figure \ref{fig:cats}), looking at groups of cats ordered by age simultaneously will provide a more compact representation of the information to the learner than learning by only looking at one pair of cats at time. Indeed, our experimental results (Section 4) indicate that GNN models provide the most accurate predictions among a set of alternative approaches.


The rest of the paper is as follows. First we provide an overview of the deployed system (Section \ref{sec:overview}) and describe the details GNN model (Section \ref{sec:approach}). Next we describe our experimental methods and results (Section \ref{sec:data}), the system as deployed (Section \ref{sec:deploy}), and related work (Section \ref{sec:rel-work}). Last, we discuss future work and our conclusions (Section \ref{sec:conclusions}).



\section{System Overview}
\label{sec:overview}
\begin{figure}
\centering
\includegraphics[width=9cm,keepaspectratio]{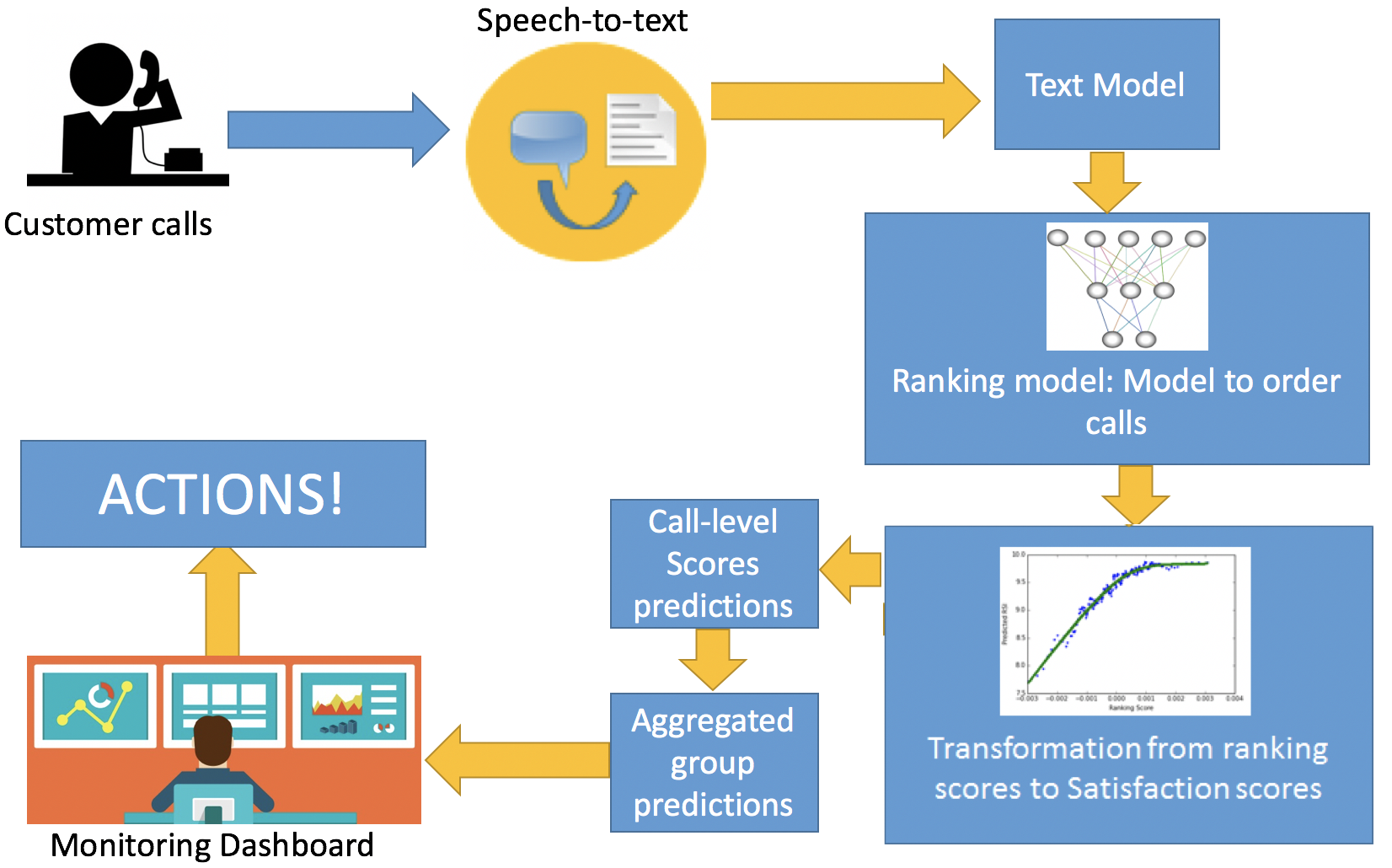}
\caption{Overview of the deployed system}
\label{fig:overview}
\centering
\end{figure}

Our company uses a third party speech-to-text software to transcribe the incoming call center calls. The files of the calls are converted into text-based-files that contain the transcription of words produced during the conversation between the caller (customer) and the customer representative. The transcription also automatically detects and maps utterances to speakers and all this extra information is provided with the transcriptions. 

The company customer care center monitors customer satisfaction by offering surveys conducted by a third party vendor to 10\% of incoming calls.  Due to low response rates, however, fewer than 1\% of all incoming calls have completed surveys. This corresponds to an average of approximately five surveys completed per month per representative. There are four topics measured by the survey: (a) If the customer felt ``valued'' during the call; (b) If the issue was resolved; (c) How polite the CR was, and (d) How clearly the CR communicated during the call. Scores range form 1 to 10 (1 lowest, 10 highest) and the four scores are averaged into an additional variable called RSI (Representative Satisfaction Index). In this paper we focus on predicting the RSI. 
 
More formally, the final modeling task is to learn a function $f(x) = \hat{y}$, mapping any given feature vector $x$ to a predicted RSI $\hat{y}$ such that on an average the difference between the predicted score and actual score $y$ is small.

Several challenges, in terms of modeling, are discovered after a quick initial inspection of the available training data:
\begin{itemize}
\item The RSI scores are highly biased towards the highest score (10), while calls with scores lower than 8 are less than 4\%. This highly skewed distribution makes building a predictive model challenging.
\item Survey scores are customer responses, thus are subjective, qualitative states heavily impacted by personal preferences.
\item The measurement scale of survey scores is ordinal; one cannot say, for example, that a score of 10 indicates double satisfaction as a score of 5. Most, if not all, standard regression techniques implicitly assume an interval or ratio scale.
\end{itemize}

Even when the final objective of the system is to predict the actual RSI scores, because of the reasons mentioned above, we propose a two-phase strategy: First we learn a model to learn a relative ranking score for the calls (ordinal ranking) and then we map the ranking score to satisfaction scores in the 1-10 scale. 
The system workflow is illustrated in Figure \ref{fig:overview} and it can be summarized by the following steps:
\begin{enumerate}
\item After a call ends, a transcript of the call is automatically produced by a speech-to-text software developed by Voci (vocitec.com). 
\item  Features are calculated for each call transcript using a word embedding model trained on more than 2 million documents including transcribed calls (for which we don't have survey scores) and other insurance-related text corpus available within the company. The resulting feature vectors are used as input features for the models described in the next step.
\item Ranking model: The ranking model is trained by sampling batch of calls and feeded into the GNN framework. The ranking loss is calculated based on the corresponding satisfaction scores. 
\item Mapping from raking scores to satisfaction scores is performed using an Isotonic Regression (IR) model \cite{dykstra1981isotonic} and thus individual (per call) satisfaction predictions are generated.
\item Aggregation of calls at the group level are stored in a database. Example groups include: per CR, per queue and per time period.
\item Aggregations are used for real-time reporting though a monitoring dashboard.
\end{enumerate}

\section{Learning rankings using GNNs}
\label{sec:approach}
Inspired by the approach proposed in \cite{meng2018efficient}, our approach is based on the fact that when ranking customer calls by satisfaction level, the learning algorithm can benefit from having access and hence exploring the similarity among multiple calls (instead of only considering pairs of calls) represented on a graph, where each node represents a call and the edges are learned based on the relationship between the to-be-learned representations of the nodes. 

Each phone call transcript is represented by an embedding vector (as described in detail in a later section) and are mapped into the graph representation. Details of our network architecture for rank learning are presented next. 

\begin{figure*}
\centering
\includegraphics[width=15cm]{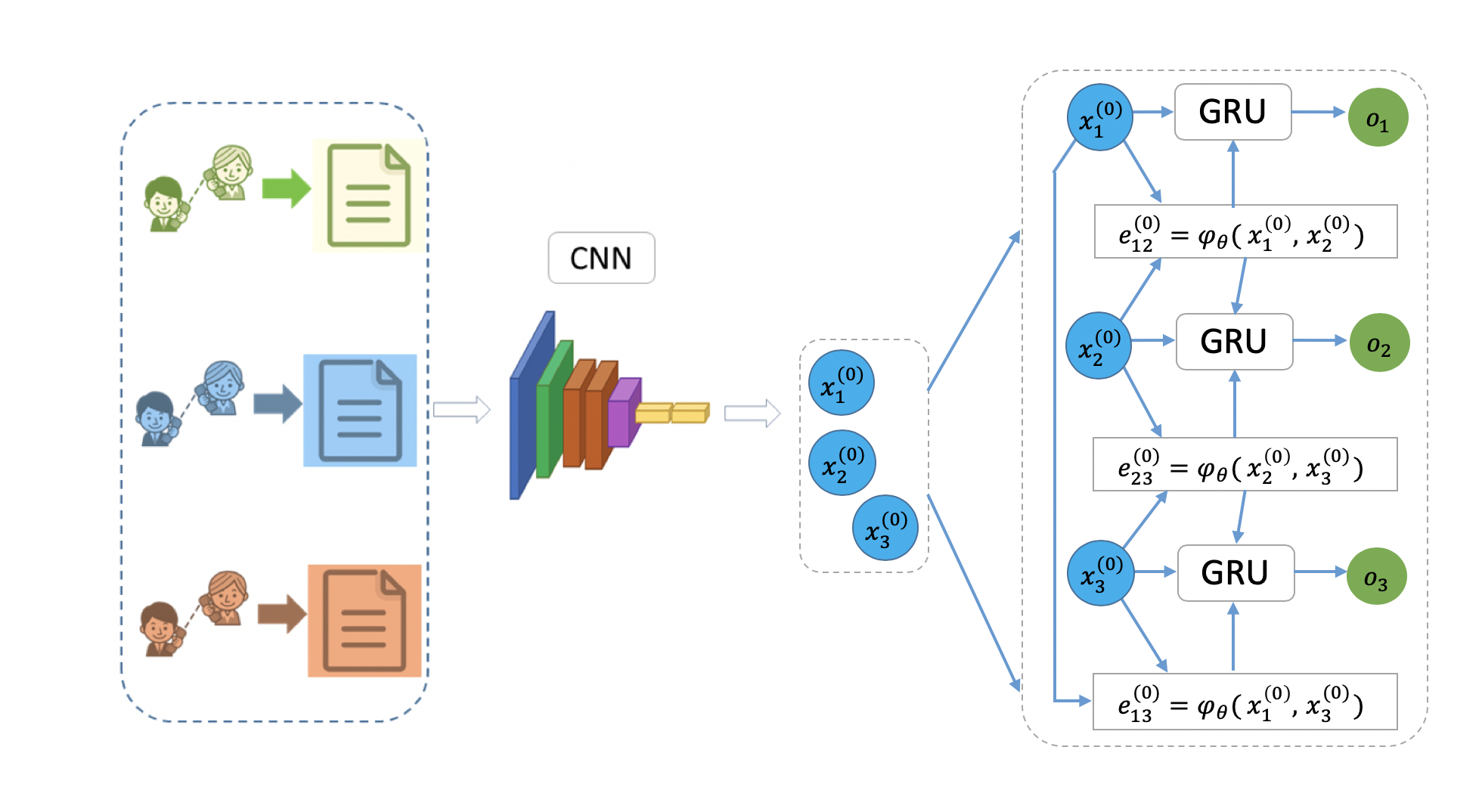}
\caption{Overview of the framework. The weights in the entire framework including those in the CNN and GNN are trained end-to-end. The edges on the graph are learned from adjacent nodes using a parametrized function ($\varphi_\vartheta$, equation. \eqref{eq:varphi}), which is shared among all edges. The messages are passed to every node from its connected nodes and edges as defined in equation. \eqref{message function}. A GRU cell combines node information and its corresponding messages generating the output. The parameters in GRU are also shared across all the nodes.}
\label{fig:framework}
\end{figure*}

\subsection{Network Architecture}
Let $\mathcal{C}=\{C_1,C_2,\cdots,C_n\}$ be the set of input calls, we assume that a set of pairwise relationship labels $\mathcal{P}_\mathcal{L}^t=\{\mathcal{L}(C_i,C_j)\}_{i,j=1;i\neq j}^n$, where $\mathcal{L}(C_i,C_j)$ indicates the relative goodness (based on survey scores) between the two calls $C_i$ and $C_j$. Given this data, a generalized GNN is trained where both the node features (embedding representations of the call transcripts) and edge weights are learned. The core architecture of our GNN is shown in Fig. \ref{fig:framework}.


As described in \cite{meng2018efficient}, we will assume that we operate on groups (or mini-batches) of a certain size (which are allowed to vary) sampled with or without replacement from the underlying training dataset. The relationships among all the calls in each mini-batch ($S$) in the training set are represented using a fully-connected graph $G_S=(V,E)$, where each node $v_i$ in $V$ corresponds to a call $C_i$ in the mini-batch $S$. For each batch $S$, the network takes in a group of calls and calculates the corresponding initial vectorial embedding representation $x_i=f(C_i)\in {\rm I\!R^n}$, where $f(\cdot)$ refers to a text embedding function. Note that in \cite{meng2018efficient}, the architecture presented considers several layers of connected Gated Recurrent Units (GRU) in a sequential manner. For this work we only consider one layer since in practice several layers did not improve model accuracy significantly, at least for our problem. Next, the network learns edge features as,
\begin{equation}
\label{learn edge}
e_{i,j}=\varphi_\vartheta\left(x_i,x_j\right),
\end{equation}
where $\varphi$ is a symmetric function parameterized with a single layer neural network:
\begin{equation}
\varphi_\vartheta(x_i,x_j) = \textrm{ReLU} \left(W\left(|x_{i}^{1}-x_{j}^{1}|,\ldots,|x_{i}^{n}-x_{j}^{n}|\right)+b\right).\label{eq:varphi}
\end{equation}
Here $x_{i}^{j}$ denotes the $j^{th}$ component of the embedding vector $x_i$. In other words, $\varphi$ is a non-linear combination of the absolute difference between the features of two nodes, $W$ and $b$ are the weight matrix and the bias respectively, and $\textrm{ReLU}(\cdot)$ is the rectified linear unit (ReLU) function.

To update the information considered at each node based on the information at the other 
nodes in the graph, we use a message function $M(\cdot)$ to 
aggregate evidence from all neighbors of each node. 
For each node $x_i$, the message is defined as below,
\begin{equation}
\label{message function}
m_i= \sum_{j,j\neq i}M\left(x_j,e_{i,j}\right).
\end{equation}

Where $M(\cdot)$ is parameterized using a single layer neural network and is defined as below, 

\begin{equation}
M(x_j, e_{i,j})=\textrm{ReLU} \left(W\left(x_j\Vert e_{i,j}\right)+b\right),
\end{equation}

Here $\Vert$ denotes the concatenation operator of two vectors, 
The parameters ($W$ and $b$) of the message function $M(\cdot)$ are shared by all nodes and edges in our graph, thus providing an explicit control on the number of parameters and making the architecture more modular, allowing us to train and perform inference with different configuration of networks according to the problem needs.

The edge features are learned as suggested by Gilmer et al. \cite{gilmer2017neural} and described in equation \eqref{learn edge}. The parameters of the edge learning function $\varphi_\vartheta$ are also shared among all nodes on the graph. Then for every node $x_i$ in the graph, a "summarized" message signal will be extracted from all the inputs and edges of this node, see \eqref{message function}. 
For each node, we use a GRU that combines the messages received from the node's neighbors and its corresponding input $x^0_i$ to produce an output through a readout function $o_i=R(x_i,m_i)$.

It is important to note that GRU units are known for their use in sequence problems. However, we are using them for a different purpose in this work. The purpose of the GRU units in this architecture is to capture and learn simultaneously the relative relations among examples in a given batch.

For our specific task of ranking calls, we use a classic ranking loss as described in \cite{burges2005learning}.  This formulation is robust to noise and is symmetric by construction so it can easily utilize batches of training data where some pairs of calls have ``equal'' scores, which is often the case here since our scores distribution is highly skewed towards 10. See Fig. \ref{fig:label_dist}. Hence, the loss function for any pair of calls $C_i$ and $C_j$ defined on the output of the graph takes the form 
\begin{equation}
\label{single_loss}
\textrm{RALLoss}=\sum_{i,j,i\neq j}-\mathcal{L}\log(P_{ij})-(1-\mathcal{L})\log(1-P_{ij}),
\end{equation}
where
\[
\mathcal{L}=
\begin{dcases}
1 & \textrm{if}~S_i \succ S_j, \\
0 & \textrm{if}~S_i \prec S_j,\\
0.5 & \textrm{otherwise},
\end{dcases}
\]
and $P_{i,j}=o_i - o_j$ (outputs of nodes $i$ and $j$). Where $S_i$ is the corresponding satisfaction score for call $C_i$.\\

This loss seeks to learn a network that, given the input calls, it simultaneously outputs pairwise labels according to the relative strength of certain attributes between each pair of calls. In this paper, we consider training a network for one attribute (RSI) at a time. However this framework can be also use to learn several tasks at the time (multi-task learning) as described in \cite{meng2018efficient}.

Our network is designed to better explore the correlated information among a batch or a group of different calls. So unlike other approaches related to relative attribute learning (including Siamese-network-based) \cite{souri2016deep,singh2016end,bockhorst2017predicting} which typically take only two training examples at a time as an input, we sample a potentially arbitrary size group of calls from the training set as input at every draw.  The size of the group need not be fixed and can be variable for learning different attributes in a single dataset or different datasets, since our network has the benefit of weight sharing on the graphical structure of the samples. 

For some of our experiments, we have used the architecture of CNN described in \cite{kim2014convolutional} with the same number of filters and static representation of the word vectors. For the initialization of the words we have used the word vectors described in Section \ref{sec:wordvecs}. The dimension of the output feature vector will be the sum of individual filter sizes multiplied by their respective feature maps.

As proposed in \cite{meng2018efficient}, we also impose a fully-connected graphical structure on the calls in each group. 

Each GRU takes the input representation for a node and it's  incoming message summary as an input, and produces the output (see Fig. \ref{fig:framework}). 
Let $x_i^0$ be the node's input representation (obtain by the CNN layer), 
$m_i$ be the message received via \eqref{message function}, and $o_i$ be the output of node through the readout function. With these notations, for our case,  the basic operations of GRU are simply given as,\\
\begin{equation}
\begin{array}{l}
\label{eqn:gru}
z_i = \sigma\left(W^z m_i + U^z x_i^0 \right), r_v = \sigma\left(W^r m_i +U^r x_i^0 \right) \mbox{ and} \\
\tilde{o}_i= \tanh\left(W m_i +U\left(r_i\odot x_i^0\right)\right),\\o_i = (1-z_i)\odot x_i^0+z_i\odot \tilde{o}_i 
\end{array}
\end{equation}

where $z$ and $r$ are the intermediate variables in the GRU cells (update and reset gates), $\sigma(x)=1/(1+e^{-x})$ is the logistic sigmoid function and $\odot$ is element-wise multiplication.

Each node in our graph maintains its internal state in the corresponding GRU, and all nodes share the same weights of the GRU, which makes our model efficient while also being able to seamlessly deal with differently sized groups as input. During the testing phase, any random number of calls are also allowed, and the network will output relative ranking scores for all of them based on the obtained value of output nodes on the graph. Note that rather than modeling a temporal process, GRUs are used here in each node to optimally combine its own input text representation with it's graph neighbor's information, hence learning in a truly non i.i.d. manner which is beneficial for learning relative rankings.  
\section{Experiments}
\label{sec:data}
\subsection{Datasets}
\subsubsection{\textbf{Call Transcripts}}
We have curated a dataset of 31000 calls to train our system out of which 20\% of calls are randomly sampled and used as a validation set. A testing set of around 8000 calls are used to measure the performance of the model. The testing set is created by making an {\it on-date} split, which means the calls in the testing set used for this work are closer to today at the time of writing this paper than the calls from the training or validation sets. We think this is important from a natural language modeling perspective since the speech-to-text software may evolve over time adding newer vocabulary or/and semantic and syntactic contexts due to new products and trends. It is important for the model to tolerate these changes and generalize efficiently, therefore we have chosen an {\it on-date } split to measure the performance of {\it Call Transcripts} in this work.

In order to deal with the potential exposure of personal identifiable information (PII), we have used regular expressions to filter out the email addresses, phone numbers, ids, people names and other characters inserted during the process of masking the calls at the call centre. In addition to this, we have used regular expressions to identify the numerical entities such as time, date, dollar signs, etc. and assigned a special token to each one of them. 

\begin{figure}
	\centering
		\includegraphics[scale=0.55]{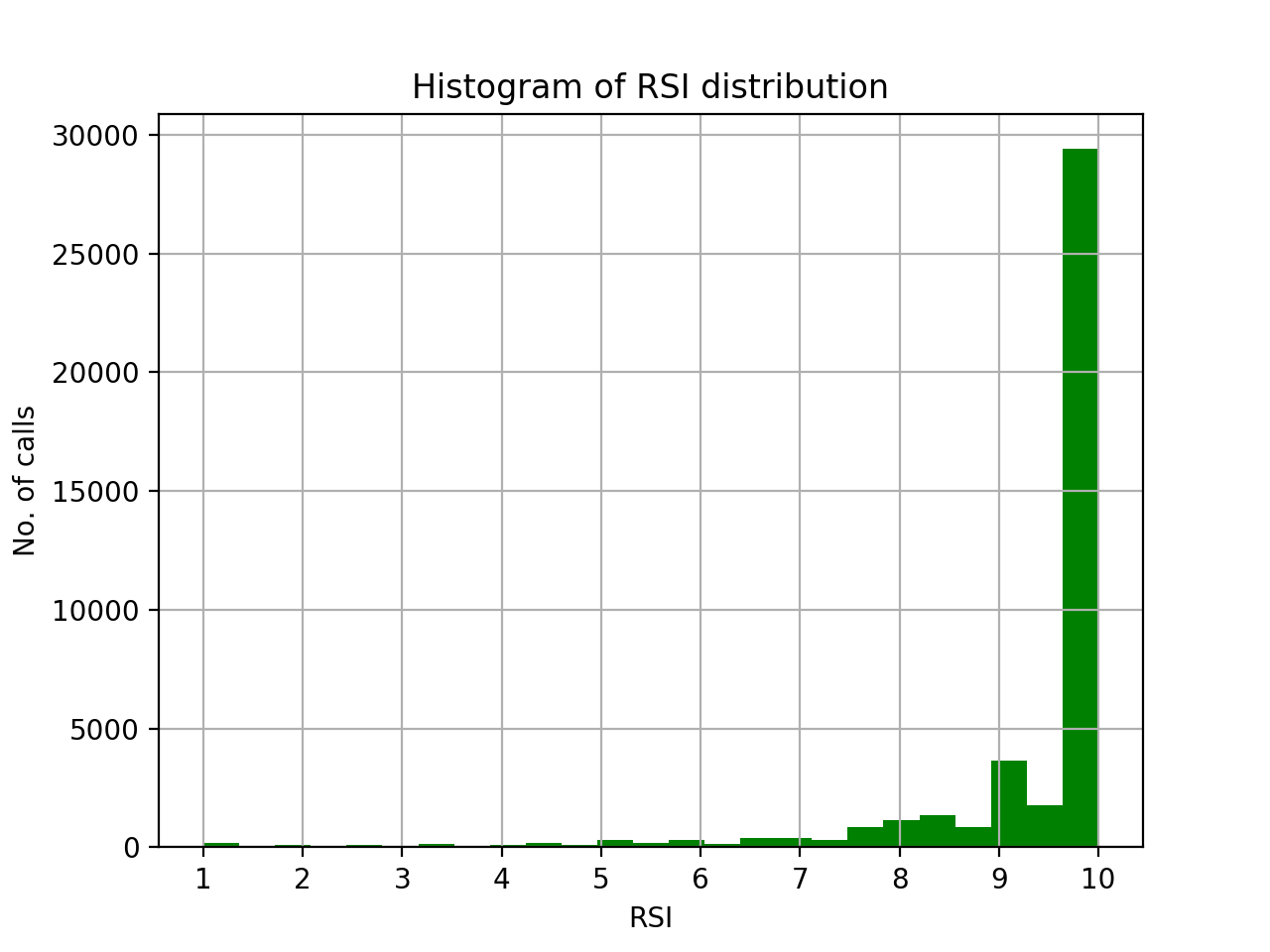}
	\caption{Histogram showing the distribution of labels}
	\label{fig:label_dist}
\end{figure}

Every month around 1\% of calls are randomly selected from the pool of calls that were received in that month as candidates for satisfaction surveys. Since only about 10\% of customers are willing to complete a survey, we end up with a limited number of surveyed calls. As mentioned before, given the ordinal nature of RSI, our modeling strategy consists of capturing the relative strength between calls during training and give out the raw rank score for each call during inference. As shown in Figure \ref{fig:label_dist}, 95\% of the labels are all ranked very high and around 90\% of calls are of class 10, thus explaining the skewed distribution of the dataset. 

\subsubsection{\textbf{Amazon Product Reviews}} To further validate the potential of the proposed approach, we have also used a publicly available dataset based on Amazon Product Reviews \cite{he2016ups}, \cite{mcauley2015image}. Specifically, we have used one of the {\it K-cores} dataset from the {\it Electronics} category and sampled around 150,000 reviews for our experiments in this work. These reviews are written by customers at Amazon and every review has an assigned rating on a scale from 1 to 5 with 1 being the most negative and 5 being the most positive customer experience.

\subsection{Features for Non-linear models}
\label{sec:wordvecs}
Word embeddings helps to represent words in a n-dimensional vector space such that words which are syntactically or semantically close to each other, are clustered together in the resulting embedded vector space.

\begin{enumerate}
    \item For the {\it Call Transcripts} dataset, we have trained a FastText model described in \cite{joulin2017bag} using 2 million transcribed calls (for which we don't have survey scores) and other text insurance-related datasets available within the company to learn a rich word embedding that captures the language commonly used in the insurance domain. Even though there are many openly available language models, none of them are trained on insurance related tasks and may not capture the insurance jargon well. Furthermore, most of them are trained on Wikipedia or News which are grammatically richer than the transcribed calls we are working with and may not be the best choice of representing them.
    \item For the {\it Amazon Product Reviews} dataset, we have used the publicly available Glove model \cite{pennington2014glove} trained on news articles.
\end{enumerate}

\subsection{Features for Linear models}
\label{sec:feat_rep}
For experiments with linear models, we have used the following feature representations:
\begin{enumerate}
\item Bag of Words + PCA (TFIDF-PCA): As described in \cite{bockhorst2017predicting}, the last quarter of the call is relevant to estimate the outcome of the satisfaction score. Therefore, we have created a composite feature vector to represent the call: first, we compute the Bag of Words features from the last quarter of the call. Then, we calculate Bag of Words features for the full call and reduced its dimension by applying PCA to it. The resulting feature vector is a concatenation of these two representations. 


\item Universal Sentence Encoder (USE): We have used the pre-trained transformer based Universal Sentence Encoder \cite{cer2018universal} which maps input to a fixed length vector representation. 

\end{enumerate}


\subsection{Models}
\label{sec:models}
We have experimented with both regression and ranking models in this work and are described as follows:

\subsubsection{\textbf{Linear models}}
\begin{enumerate}
    \item Lasso: This is a regression based method with $L_1$ regularization \cite{tibshirani1996regression}. We have chosen this model because of its robustness with respect to outliers. 

    \item Rank Score(RS): This is a linear ranking model described in \cite{bockhorst2017predicting}. The authors in their work describe that the Isotonic Regression (IR) is used to map rank scores to the actual scores. In this work, we are omitting the IR part of the model as our reported measures (described in Section \ref{sec:metric}) doesn't depend on this mapping.
\end{enumerate}

\subsubsection{\textbf{Non-linear models}}
\begin{enumerate}
    \item CNN: This is a regression based model. Inspired from the architecture proposed in \cite{kim2014convolutional}, we have chosen this model and have attached a linear combination layer at the end of the pooling layer and minimized the mean-squared-error (MSE) during training.
    \item CNN-GNN: This is our proposed ranking based model and is described in Section \ref{sec:approach}.
\end{enumerate}

\subsection{Evaluation metric}
\label{sec:metric}

\subsubsection{\textbf{Spearman}} To test the performance of models, one of the metrics we have chose is the Spearman correlation to measure the association between observed and predicted scores. Spearman correlation measures the strength and direction of the monotonic association between the two variables and its often used to evaluate performance of ranking algorithms \cite{zimmermann2007predicting}. The Spearman coefficient is a number between $-1$ to $+1$, where $+1$ signifies perfect correlation, $-1$ signifies perfectly opposite correlation and 0 indicates no correlation at all.

\subsubsection{\textbf{Precision@k}} We also report the Precision@k \cite{jarvelin2017ir} which is the proportion of items in the top-k set that are true positives. This is an important metric for our call center since resources are limited and usually there is only bandwidth to explore a low percentage of the calls made in a certain period of time.

\subsection{Experimental Settings}
We have used a batch size of $64$, number of nodes in the graph as $5$ (for CNN-GNN) and a learning rate of $\alpha=10^{-5}$. We used the Adam optimizer \cite{kingma2014adam} with $\beta_1$ set to 0.9 and $\beta_2$ set to 0.999 and initialized the weights of the network with Xavier initialization \cite{glorot2010understanding}. For the non-linear models,  we have used a dropout of 50\% after the max-pooling layer of the CNN and for the CNN-GNN model, we have used another dropout of 50\% at each GRU cell of the graph. We have used $L_2$ regularization of 0.5 in all of the experiments and the parameter is chosen based on a grid search using the validation set. All the experiments were conducted on a K80 GPU Amazon Web Services (AWS) instance. 

To select the number of nodes for this CNN-GNN model, we ran several experiments varying the numbers from 2 to 10. 


\subsection{Inference}
\label{sec:inference}
For most of existing graph-based deep learning methods, inference is mostly done in a transductive manner \cite{gnnsurvey}. For our GNN-based approach inference involves calculating the edges weights, aggregating messages received from each neighbor and passing this aggregated message to the GRU (see equation \ref{eqn:gru}). Hence, Inference for each testing example depends on all the other examples that are simultaneously considered in the graph when scoring. However, for our case we want to be able to generate independent absolute (not relative) ranking scores for each testing sample in an inductive manner.  In order to achieve this we use the concept of ``anchors" borrowed from the kernel classifiers literature \cite{anchors}. 
The idea is simple: to select a small group of training point as fixed reference points to compare against when testing in a batch. Hence, any new unseen testing point $x$ can be scored by the final trained GNN model $f$  by calculating $\tilde{f}(x)=f(x,\tilde{x_1},\ldots,\tilde{x_k})$ where $(\tilde{x_1},\ldots,\tilde{x_k})$ is the set of prefixed $k$ anchors.


\subsection{Results}

 We evaluate the Spearman correlation for all the models described in \ref{sec:models}. See Table \ref{tab:spearman_table}. The proposed CNN-GNN model outperforms all the other models for the two datasets we have used in this work. In Table \ref{tab:amzn_table}, we report the precision@k for different values of $k$ corresponding to top percentile of the ranked predictions for both datasets. 
 
 For the Amazon product reviews dataset, we have chosen the reviews with a rating of 1 as the relevant items and all the other ratings as non-relevant. The proposed CNN-GNN model outperforms all the other compared models. 
 
 Similarly, in Table \ref{tab:transcripts_table}, we report the precision@k of items with a lower RSI scores (lower than 7). Again, the proposed model outperforms the other algorithms specially at percentiles close to the top of the ranked prediction list.

	\label{tab:rank_results}

\begin{table*}[t]
\caption{Table reporting the Spearman Correlation (x100). The models are described in Section \ref{sec:models}. In RS and Lasso, we use TFIDF-PCA features and we use USE features in RS$^\star$ and Lasso$^\star$ }
\label{tab:spearman_table}
\centering
{\small
\begin{tabular}{p{3.5cm}p{2cm}p{2cm}p{2cm}p{2cm}p{2cm}p{2cm}}
\toprule
Dataset & CNN-GNN & RS & RS$^\star$ & Lasso & Lasso$^\star$ & CNN \\
\midrule
Call Transcripts & \textbf{29.85} & 26.21 & 17.76 & 26.55 & 24.55 & 26.74  \\
\midrule
Amazon Product Reviews & \textbf{71.25}  & 67.44 & 50.49 & 68.37 & 67.34 & 67.19 \\
\bottomrule
\end{tabular}
}
\end{table*}


\begin{table*}[t]
\caption{Table reporting the Precision@k for the Amazon Product Reviews dataset where k corresponds to the top k percentage of the ranked predictions. In order to calculate the metric, we have chosen the reviews with a rating of 1 as the relevant items and all the other ratings as non-relevant. The models are described in Section \ref{sec:models}. In RS and Lasso, we use TFIDF-PCA features and we use USE features in RS$^\star$ and Lasso$^\star$ }
\label{tab:amzn_table}

\centering
{\small
\begin{tabular}{p{2cm}p{2cm}p{2cm}p{2cm}p{2cm}p{2cm}p{2cm}}
\toprule
Precision@k & {\bf CNN-GNN} & RS & RS$^\star$ & Lasso & Lasso$^\star$ & CNN \\
\midrule
1\% & {\bf 0.873} & 0.815 & 0.707 & 0.847 & 0.794 &0.747  \\
2\% & {\bf 0.834} & 0.776 & 0.696 & 0.794 & 0.771 &0.734  \\
3\% & {\bf 0.803} & 0.753 & 0.682 & 0.762 & 0.749 &0.735  \\
4\% & {\bf 0.774} & 0.725 & 0.666 & 0.743 & 0.733 &0.709  \\
5\% & {\bf 0.755} & 0.702 & 0.657 & 0.715 & 0.717 &0.698  \\
10\% & {\bf 0.687}  & 0.640 & 0.588 & 0.651 & 0.653 &0.622 \\
25\% & {\bf 0.526} & 0.506 & 0.475 & 0.512 & 0.510 &0.504 \\
50\% & 0.366 & 0.364 & 0.347 & 0.365 & 0.364 & \textbf{0.367} \\
75\% & 0.269 & 0.269 & 0.263 & 0.269 & 0.268 &0.269 \\
100\% & 0.205 & 0.205 & 0.205 & 0.205 & 0.204 &0.205\\
\bottomrule
\end{tabular}
}
\end{table*}


\begin{table*}[t]
\caption{Table reporting the Precision@k for the Call Transcripts dataset where k corresponds to the top k percentage of the ranked predictions. In order to calculate the metric, we have chosen the calls with RSI $\leq$7 as relevant items and all the others as non-relevant. The models are described in Section \ref{sec:models}. In RS and Lasso, we use TFIDF-PCA features and we use USE features in RS$^\star$ and Lasso$^\star$ }
\label{tab:transcripts_table}

\centering
{\small
\begin{tabular}{p{2cm}p{2cm}p{2cm}p{2cm}p{2cm}p{2cm}p{2cm}}
\toprule
Precision@k & CNN-GNN & RS & RS$^\star$ & Lasso & Lasso$^\star$ & CNN \\
\midrule
1\% & \textbf{0.476} & 0.333 & 0.238 & 0.357 & 0.404 & 0.404  \\
2\% & \textbf{0.380} & 0.273 & 0.178 & 0.309 & 0.297 & 0.333  \\
3\% & \textbf{0.373} & 0.277 & 0.158 & 0.333 & 0.317 & 0.309  \\
4\% & \textbf{0.380} & 0.291 & 0.172 & 0.351 & 0.315 & 0.297  \\
5\% & \textbf{0.338} & 0.30  & 0.190 & 0.328 & 0.309 & 0.261  \\
10\% & 0.257 & 0.245 & 0.164 & \textbf{0.269} & 0.238 & 0.211  \\
25\% & 0.170 & 0.170 & 0.127 & \textbf{0.179} & 0.162 & 0.165  \\
50\% & 0.118 & 0.117 & 0.103 & \textbf{0.119} & 0.113 & 0.114  \\
75\% & 0.09 & 0.09 & 0.08 & 0.09 & 0.09 & 0.09  \\
100\% & 0.07 & 0.07 & 0.07 & 0.07 & 0.07 & 0.07  \\
\bottomrule
\end{tabular}
}
\end{table*}
\section{Brief overview of the deployment Architecture}
\label{sec:deploy}
When a customer calls, our system records the audio from both speakers and store them in our data-warehouse ({\it I3 call center}). Then, the audio files are processed in the following way: First, they are transcribed into text using a speech-to-text software described in Section \ref{sec:overview}. Second, these transcriptions are processed through a sentiment analysis and emotion engine to extract extra insights from the data. We also run these transcripts through a set of business rules to assign different tags releveant to the business like: reason of the call, call topics, etc. All of this newly generated and call-derived meta-data, along with the transcription of the calls are written to a JSON file and sent to our internal Hadoop eco-system for storage. 

\begin{figure*}
\centering
	\includegraphics[scale=0.35]{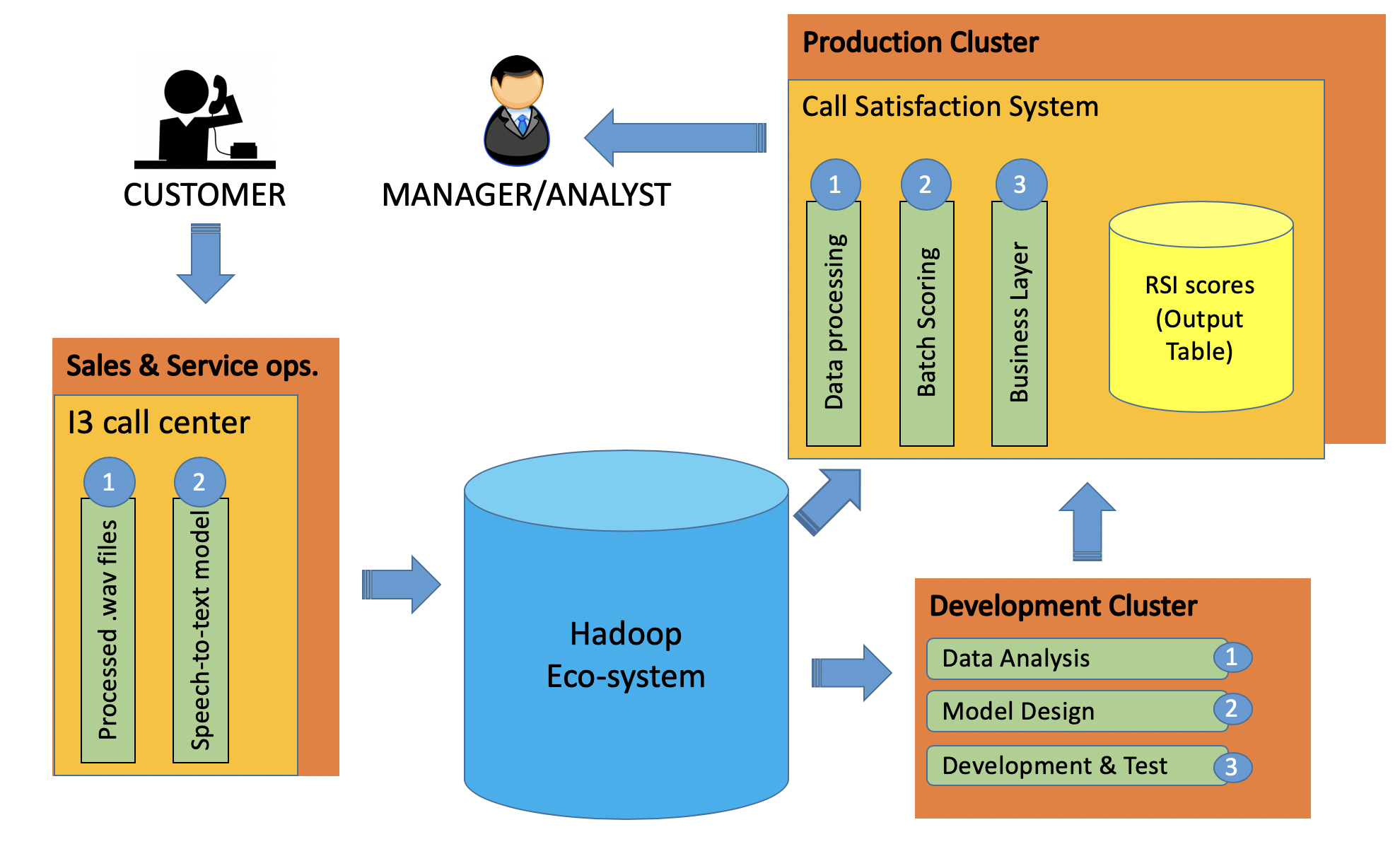}
	\caption{Architecture overview of our deployed system}
	\label{fig:architechture}
\end{figure*}

The call transcripts are then used as input to our model to predict RSI. The other meta-data is used for other business-related insights. As shown in Figure \ref{fig:architechture}, we have development and production clusters. The development cluster is used to analyze and create insights from the data as well as to train and create predictive models. The models that are created in this cluster are tested and then deployed to a production cluster for daily scoring. Once a day, a batch script is run to score all the calls that were received that day. The resulting predicted scores are stored in a database for the business to consume them. These generated predictions are used for different purposes as: agent education/feedback, process improvement, and other use-cases that help the company to serve its customers more efficiently. For many of this tasks being able to identify low scoring calls with high precision is paramount to optimize utilization of resources and managers' time.

\section{Related Work}
\label{sec:rel-work}
Research studies on emotion recognition using human-human real-life corpus extracted from call center calls are limited. In~\cite{Vau12}  a system for emotion recognition in the call center domain is proposed. The goal was to classify parts of dialogs into different emotional states. 

Park and Gates~\cite{Par09} developed a method to automatically measure customer satisfaction by analyzing call transcripts in near real-time. They built machine learning models that predict the degree of customer satisfaction in a scale from 1 to 5 with an accuracy of 66\%. Sun \textit{et al}.~\cite{Sun16} adopted a different approach, based on fusion techniques, to predict the user emotional state from dialogs extracted from a Chinese Mobile call center corpus. 

Recently, convolutional neural networks have been used on raw audio signals for prediction of self-reported customer satisfaction from call center\cite{Seg16}. They pretraining a network on debates from TV shows . Then, the last layers of the network are fine-tuned with more than 18000 conversations from several call centers. The CNN-based system achieved comparable performance to the systems based on traditional hand-designed features. 

Emotion recognition in the call center domain usually involves rating based on an ordinal scale.  Indeed, psychometric studies show that human ratings of emotion do not follow an absolute scale~\cite{Meta13}.


There are several algorithms which specifically benefit from the ordering information and yield better performance than nominal classification and regression approaches. There are algorithms that focus on comparing training examples in a pairwise manner using binary linear classifiers \cite{Herb99,Harp02}. Crammer and Singer~\cite{Cramm05} developed an ordinal ranking algorithm based on the online perceptron algorithm with multiple thresholds. 

Some areas where ordinal ranking problems are found include medical research~\cite{Pere14}, brain computer interface~\cite{Yoon11}, credit rating~\cite{Kim12}, facial beauty assessment~\cite{Yan14}, image classification~\cite{Tian14}, and more. All these works are examples of applications of  ordinal ranking models, where exploiting ordering information improves their performance with respect to their nominal counterparts. 

The concept of graph neural network (GNN) was first
proposed in \cite{gnn}, which extended existing neural networks for learning with data and domains that benefit for being represented in graph domains. Generally, in a GNN, each node is influenced by its features and
the related nodes to it (neighbors). This framework fits naturally to the problem of ordinal ranking where the ordering relations can be represented by edges in a graph. There are prior applications of GNN to the ranking problem. For example in \cite{Rankingattack} a GNN is used to rank attacks in network security setting. A comprehensive very recent survey on GNNs and their applications is presented in \cite{gnnsurvey}.

\section{Conclusions}
\label{sec:conclusions}
This paper describes an efficient updated system for predicting self-reported satisfaction scores of customer phone calls. Our system has been implemented into a production pipeline that is currently predicting caller satisfaction of approximately  30,000  incoming  calls  each  business  day  and  generating  frequent reports read by call-center managers and decision makers in our company to potentially improve processes that impact daily customer interactions. 

In this work, we described a newly implemented learning algorithm based on GNNs that has considerably improved our system accuracy and that can be generalized to other ranking tasks inside and outside the customer satisfaction and the insurance domain. 

We  presented  empirical  evaluation based on a large number of real customers calls that shows that this approach yields  more  accurate  satisfaction  predictions  than  standard  regression and ranking  models while only using raw call speech-to-text transcriptions.

We are excited to report that post-deployment validation of our system indicates that the average satisfaction prediction for groups of calls agrees very strongly with actual satisfaction scores, especially for large groups.

As future work, We are interested in human-in-the-loop / active learning approaches where a feedback loop can be used to improve the system as it is used. Under this paradigm,  restricting supervision to paired comparisons can be tedious and wasteful due to the quadratic nature of possible comparisons. Instead we could present the oracle with the task of ordering a small subset of examples at the time, which could result in a more streamlined experience for the labeler and more efficient way to capture feedback.

\bibliographystyle{unsrt}  

\bibliography{acmart.bib}

\end{document}